\documentclass[letterpaper]{article} 
\usepackage{aaai2026}  
\usepackage{times}  
\usepackage{helvet}  
\usepackage{courier}  
\usepackage[hyphens]{url}  
\usepackage{graphicx} 
\usepackage{natbib}  
\usepackage{caption} 
\usepackage{algorithm}

\usepackage[utf8]{inputenc} 
\usepackage[T1]{fontenc}    
\usepackage{url}            
\usepackage{booktabs}       
\usepackage{amsfonts}       
\usepackage{nicefrac}       
\usepackage{microtype}      
\usepackage{xcolor}         
\usepackage{graphicx} 
\usepackage{amsmath} 

\usepackage{algorithm}
\usepackage{algpseudocode}
\usepackage{amssymb}
\usepackage{multirow} 
\usepackage{colortbl}
\usepackage{threeparttable}
\usepackage{booktabs}   

\usepackage{newfloat}
\usepackage{listings}
\DeclareCaptionStyle{ruled}{labelfont=normalfont,labelsep=colon,strut=off} 
\floatstyle{ruled}
\newfloat{listing}{tb}{lst}{}
\floatname{listing}{Listing}
\title{Internal Representations as Indicators of Hallucinations in Agent Tool Selection}
\author{
    Kait Healy\equalcontrib,
    Bharathi Srinivasan \equalcontrib,
    Visakh Madathil\equalcontrib,
    Jing Wu \equalcontrib
}
\affiliations{
    Amazon\\
    410 Terry Avenue North, Seattle, WA 98109, United States

}

\usepackage{bibentry}

\begin{document}

\maketitle

\begin{abstract}
Large Language Models (LLMs) have shown remarkable capabilities in tool calling and tool usage, but suffer from hallucinations where they choose incorrect tools, provide malformed parameters and exhibit 'tool bypass' behavior by performing simulations and generating outputs instead of invoking specialized tools or external systems. This undermines the reliability of LLM based agents in production systems as it leads to inconsistent results,  and bypasses security and audit controls. Such hallucinations in agent tool selection require early detection and error handling. Unlike existing hallucination detection methods that require multiple forward passes or external validation, we present a computationally efficient framework that detects tool-calling hallucinations in real-time by leveraging LLMs' internal representations during the same forward pass used for generation. We evaluate this approach on reasoning tasks across multiple domains, demonstrating strong detection performance (up to 86.4\% accuracy) while maintaining real-time inference capabilities with minimal computational overhead, particularly excelling at detecting parameter-level hallucinations and inappropriate tool selections, critical for reliable agent deployment.
\end{abstract}

\section{Introduction}
Large Language Models (LLMs) have demonstrated remarkable capabilities as autonomous agents, enabling them to interact with external APIs, execute code, and orchestrate complex workflows~\cite{brown2020language,schick2023toolformer,qin2024toolllm}. However, these models exhibit a critical vulnerability: \textbf{tool-calling hallucinations}, where they generate structurally plausible but functionally incorrect tool calls. Unlike textual hallucinations, tool-calling hallucinations manifest as inappropriate tool selection, malformed parameters, incorrect tool chaining, tool bypass behavior, or semantically incorrect function invocations that can lead to system failures or data corruption.

The detection of tool-calling hallucinations presents unique challenges. Tool calls possess rigid structural and semantic constraints—parameters must conform to specific types and must be accurately interpreted from the query to the agent, required arguments cannot be omitted, and function names must exist in the available repertoire. Moreover, the consequences of undetected hallucinations can be severe in computational domains where precision is critical. Incorrect tool usage can also exacerbate security vulnerabilities of agents due to incorrect data access

To address these challenges, we propose a novel approach that leverages the internal representations of LLMs during tool call generation to detect hallucinations in real-time. Our method builds upon recent advances in understanding LLM internal states~\cite{su2024mind,azaria2023internal} and extends them to structured tool calling. We introduce an unsupervised training framework that automatically generates labeled data by masking ground-truth tool calls, prompting LLMs to predict appropriate functions, and training lightweight classifiers on the resulting contextualized embeddings.

Our key contributions are: 
\begin{itemize}
    
    \item We demonstrate that internal representations of LLMs contain discriminative information for detecting tool-calling hallucinations in reasoning tasks.
    
    \item We evaluate our approach on tool calling scenarios, showing effective real-time detection capabilities with minimal computational overhead.
\end{itemize}

\section{Related Work}
\subsection{Hallucination Detection in Large Language Models}
Hallucination detection has emerged as a critical challenge in deploying LLMs for real-world applications. Early approaches focused on post-processing methods that analyze generated text for factual inconsistencies~\cite{maynez2020faithfulness,zhou2020detecting,ji2023survey, wu2025unfixing, wu2025building,wu2023hallucination}. These methods typically require external knowledge sources or multiple model generations to assess consistency.
Recent work has explored uncertainty-based approaches that leverage model confidence signals. \citet{kadavath2022language} demonstrated that LLMs can express uncertainty about their knowledge, while \citet{zhang2023enhanced} proposed enhanced uncertainty estimation methods for hallucination detection. However, these approaches struggle with the discrete nature of tool calling, where traditional uncertainty measures may not capture semantic correctness of API usage.
Consistency-based methods represent another major direction. SelfCheckGPT~\cite{manakul2023selfcheckgpt} generates multiple responses and checks for consistency, assuming that hallucinated content will be inconsistent across samples. Non Contradiction Probability (NCP)~\cite{hou2025probabilistic} and Semantic Similarity~\cite{kuhn2023semantic} are two techniques using this principle of consistency to detect hallucinations. After sampling, NCP probabilistically scores whether a response contradicts entries in a curated belief set using information-theoretic scoring or belief tree propagation, while semantic similarity measures the degree of meaning alignment between a model’s outputs from multiple samples using cosine similarity or nearest-neighbor score between them. \citet{li2023halueval} introduced evaluation frameworks for assessing hallucination detection methods. While effective for free-form text, these approaches face limitations in tool calling scenarios where there may be unique correct solutions, making consistency-based detection less reliable.
\subsection{Tool-Augmented Language Models}
The integration of external tools with language models has gained significant attention as a means to extend LLM capabilities beyond the parametric knowledge acquired from their training data. Toolformer~\cite{schick2023toolformer} pioneered self-supervised learning for tool use, demonstrating that LLMs can learn when and how to call APIs through minimal demonstrations. This work established the foundation for automated tool learning but did not address the detection of incorrect tool usage.
ToolLLM~\cite{qin2024toolllm} scaled tool learning to thousands of real-world APIs, introducing comprehensive training frameworks and evaluation benchmarks. Gorilla~\cite{patil2024gorilla} focused specifically on API calling accuracy, achieving strong performance through retrieval-augmented training. ReAct~\cite{yao2023react} proposed reasoning and acting paradigms that interleave tool use with reasoning, while \citet{lu2023chameleon} developed plug-and-play compositional reasoning systems.
Despite these advances, existing work primarily focuses on improving tool-use accuracy during training rather than detecting errors during inference. Our work addresses this gap by providing real-time detection capabilities that can identify when models generate inappropriate tool calls.
\subsection{Internal Representation Analysis}
Understanding the internal mechanisms of neural networks has become increasingly important for building reliable AI systems. \citet{azaria2023internal} demonstrated that internal states of LLMs indicate the veracity of responses, showing that classifiers trained on hidden representations can detect when models produce false statements. This work provided crucial evidence that internal representations encode semantic properties beyond surface-level text generation.
Building on this foundation, \citet{su2024mind} developed unsupervised methods for real-time hallucination detection using contextualized embeddings from transformer layers. Their approach eliminated the need for manual annotation by automatically generating training data through entity masking in Wikipedia articles. Our work extends these insights to the structured domain of tool calling, where the challenges and patterns differ significantly from free-form text generation.
Recent work has also explored mechanistic interpretability of transformers~\cite{elhage2021mathematical,wang2022interpretability}, providing insights into how these models process and represent information. \citet{meng2022locating} investigated knowledge storage and editing in transformers, while \citet{li2023inference} analyzed attention patterns in reasoning tasks. These insights inform our understanding of where tool-calling information might be encoded within transformer representations.
\subsection{Agent Systems and Reliability}
The deployment of LLM-based agents in real-world scenarios has highlighted the need for robust error detection and mitigation strategies. \citet{xi2023rise} surveyed the landscape of autonomous agents, identifying reliability as a key challenge for practical deployment. \citet{wang2023voyager} demonstrated lifelong learning agents but noted the challenges of ensuring consistent performance across diverse environments.
Recent work has explored agent evaluation frameworks. \citet{liu2023agentbench} introduced comprehensive benchmarks for agent capabilities, while \citet{xu2023lemur} proposed evaluation metrics for tool-using agents. However, these evaluation frameworks primarily focus on task completion rather than real-time error detection during agent execution.
Safety and alignment in agent systems have also received attention. \citet{hendrycks2023overview} discussed risks associated with autonomous AI systems, while \citet{kenton2021alignment} explored alignment challenges specific to agent deployment. Our work contributes to this area by providing mechanisms for detecting potentially harmful or incorrect agent actions before they impact external systems.
\subsection{Evaluation Methodologies for tool calling}
Benchmarking tool calling capabilities has become increasingly sophisticated as the field has matured. APIBench~\cite{patil2024gorilla} established evaluation protocols for API calling accuracy, focusing on syntactic and semantic correctness of generated calls. ToolBench~\cite{qin2024toolllm} expanded evaluation to include multi-step reasoning and complex tool interactions.
Recent work has emphasized the importance of real-world evaluation scenarios. \citet{tang2023toolalpaca} introduced evaluation datasets based on actual API usage patterns, while \citet{ruan2023tptu} explored evaluation in dynamic environments where API specifications change over time. These works highlight the challenges of tool calling evaluation but have not addressed real-time error detection during generation.

\begin{figure*}[t!]
\begin{center}
    \includegraphics[width=0.95\linewidth]{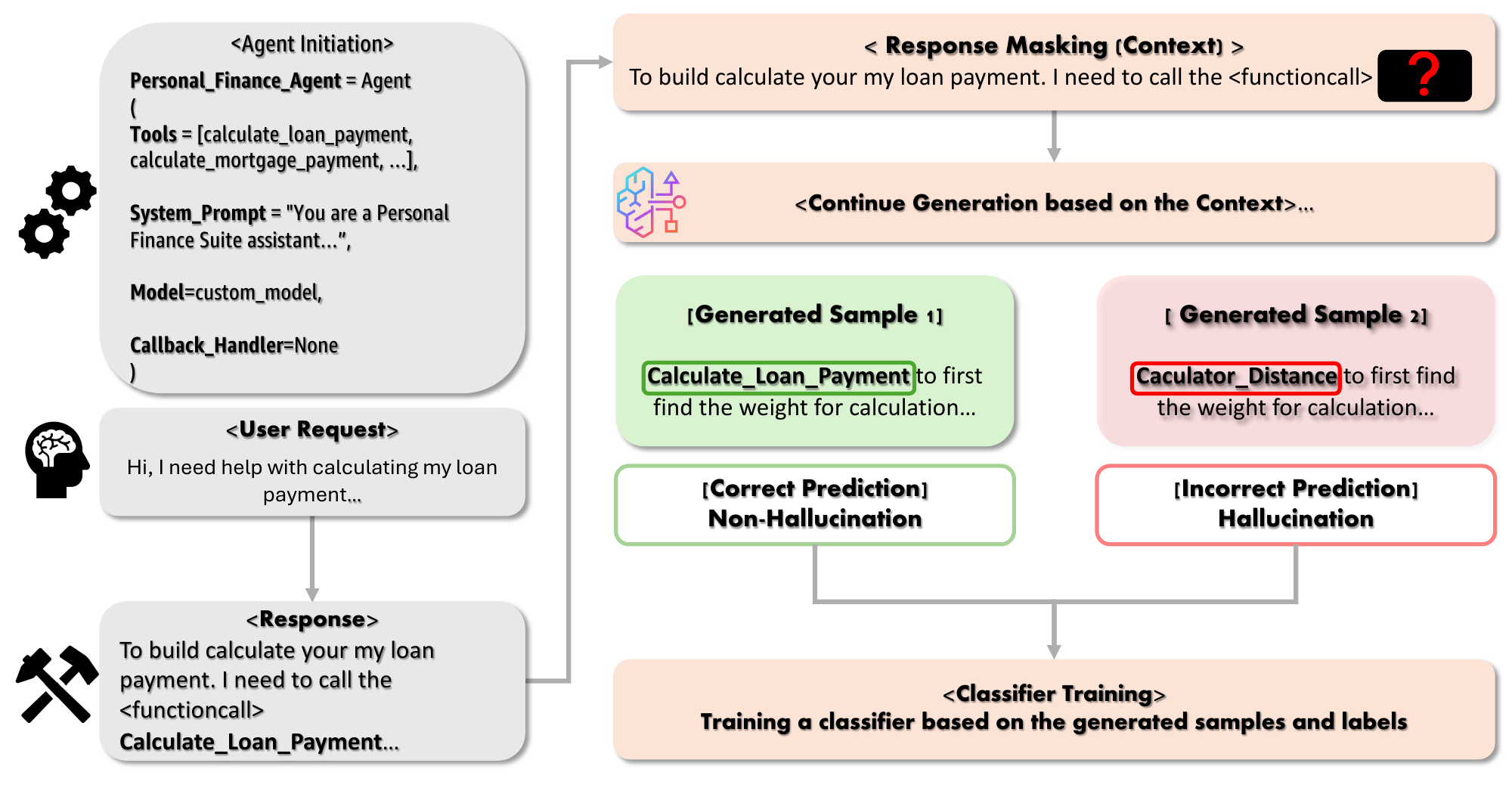}
\end{center}
\vspace{-3mm}
  \caption{\textbf{Unsupervised, single-pass training pipeline for tool-calling hallucination detection.} Given a user query and available tools, the LLM generates candidate responses containing tool calls. We \emph{mask} the tool-call segment to create a context-only prompt, ask the model to predict the call again, and assign labels by agreement with the reference call (non-hallucination vs.\ hallucination). We then train a lightweight classifier on last-layer representations to distinguish correct from hallucinated calls; at inference, the classifier scores each proposed call in real time and gates execution accordingly. 
}
\vspace{-4mm}
\label{fig2: interations}
\end{figure*}
\section{Problem Formulation}

We formalize tool-calling hallucination detection as a binary classification task operating on the internal representations of the last layer of large language models (LLMs) during tool call generation. Let $\mathcal{F} = \{f_1, f_2, \ldots, f_n\}$ denote the set of available functions in an agent's toolkit. Given a user query $q$ and contextual information $c$, an LLM $\mathcal{M}$ generates a tool call $\tilde{f}(\boldsymbol{a})$ where $\tilde{f} \in \mathcal{F} \cup \{\emptyset\}$ and $\boldsymbol{a}$ represents the function arguments.

We define a tool-calling hallucination as occurring when any of the following conditions hold:
\begin{itemize}
\item[(1)] \textbf{Function Selection Error}: $\tilde{f} \notin \mathcal{F}$ (invoking a non-existent function)
\item[(2)] \textbf{Function Appropriateness Error}: $\tilde{f} \in \mathcal{F}$ but $\tilde{f}$ is semantically inappropriate for query $q$ given context $c$
\item[(3)] \textbf{Parameter Error}: $\boldsymbol{a} \notin \mathcal{D}_{\tilde{f}}$, where $\mathcal{D}_{\tilde{f}}$ denotes the valid argument domain for function $\tilde{f}$
\item[(4)] \textbf{Completeness Error}: Required parameters are missing from $\boldsymbol{a}$
\item[(5)] \textbf{Tool Bypass Error}: Generating the output without using $\tilde{f} \in \mathcal{F}$
\end{itemize}

Our objective is to learn a classifier $h_\theta: \mathbb{R}^d \rightarrow \{0, 1\}$ that maps the internal representation $\mathbf{z} \in \mathbb{R}^d$ of the LLM at tool call generation time to a binary decision, where 1 indicates a hallucinated tool call and 0 indicates a correct tool call. While we categorize hallucinations into five distinct types for theoretical completeness, our current work focuses on unified binary detection across all error categories, enabling the model to learn general patterns of incorrectness while maintaining practical deployment simplicity.

\section{Method}\label{sec:method}

We detect hallucinations in AI agent tool calling based on the hypothesis that the representations from the final transformer layer during tool call generation contain sufficient information to distinguish between correct and hallucinated calls.

\paragraph{Setup and Notation}
Let $\mathcal{D}=\{(q_i, c_i, f_i^*, \boldsymbol{a}_i^*, y_i)\}_{i=1}^N$ denote our collected tool calling instances, where $q_i$ represents the user query, $c_i$ the contextual information, $f_i^*$ the ground-truth function, $\boldsymbol{a}_i^*$ the ground-truth arguments, and $y_i \in \{0,1\}$ the binary label indicating whether the predicted tool call constitutes a hallucination. Given input $(q_i, c_i)$, the LLM $\mathcal{M}$ generates a predicted call $\tilde{f}_i(\boldsymbol{a}_i)$. We extract features from the final layer hidden states of the same forward pass that produces this prediction.

\paragraph{Data Collection and Label Generation}\label{sec:data-collection}
We utilize the Glaive Function-Calling dataset \cite{glaiveai_glaive_function_calling_v2_2024} from Hugging Face, which comprises multi-domain agent interactions annotated with structured tool usage information and corresponding parameters, covering domains such as personal health, finance, and general utility calculations. We pre-process this dataset by grouping samples based on their designated tool names and normalizing argument structures to ensure consistency in both function and parameter representation. To evaluate tool-specific behaviors and hallucination rates, we develop specialized agents using the Strands framework \cite{strands_agents_2025} that are programmatically matched to tool names and parameter schemas extracted from the preprocessed samples. These agents are then deployed to receive data samples filtered by tool assignment, enabling precise measurement of agent performance and tool invocation fidelity under controlled experimental conditions.

We collect agent interaction data from various reasoning domains spanning personal health calculations, financial computations, sustainability metrics, and general calculator functions—domains where tool calling precision is critical for AI system accuracy. For each interaction $(q_i, c_i, f_i^*, \boldsymbol{a}_i^*)$, we obtain training labels through the following procedure: (1) remove the reference call $(f_i^*, \boldsymbol{a}_i^*)$ from the prompt while preserving $(q_i, c_i)$, (2) prompt $\mathcal{M}$ to obtain agent response $\tilde{f}_i(\boldsymbol{a}_i)$, extract tool call if present and cache final-layer states from this forward pass, and (3) assign binary labels by comparing predicted and reference calls through function name matching and argument canonicalization. Specifically, a prediction receives label $y_i = 1$ (hallucinated) if the function name differs from the reference or if arguments fail to match after normalization; otherwise $y_i = 0$ (correct). This approach yields naturally occurring hallucination examples in computational domains where mathematical precision is paramount.

\paragraph{Final-Layer Feature Extraction}\label{sec:lastlayer}
We extract features from the final transformer layer hidden states of the forward pass that generated $\tilde{f}_i(\boldsymbol{a}_i)$. Let $\mathbf{h}^{(L)}_t \in \mathbb{R}^d$ denote the final-layer hidden state at token position $t$, where $L$ is the number of transformer layers. We identify three critical token positions within the structured tool call: (1) $t^{\text{func}}$: the initial sub-token of the predicted function name, (2) $\mathcal{T}_{\text{args}} = \{t_1, t_2, \ldots, t_k\}$: all tokens spanning the argument region, and (3) $t^{\text{end}}$: the closing delimiter token (e.g., ``)'' or ``\}'').

We construct a compact feature representation by concatenating representations from these three semantic regions:
\begin{equation}
\mathbf{z}_i = \Pi\left(\mathbf{h}^{(L)}_{t^{\text{func}}} \,\|\, \frac{1}{|\mathcal{T}_{\text{args}}|}\sum_{t \in \mathcal{T}_{\text{args}}} \mathbf{h}^{(L)}_{t} \,\|\, \mathbf{h}^{(L)}_{t^{\text{end}}}\right) \in \mathbb{R}^m
\end{equation}
where $\|$ denotes vector concatenation and $\Pi: \mathbb{R}^{3d} \rightarrow \mathbb{R}^m$ represents an optional linear projection (identity mapping by default, yielding $m = 3d$).

\paragraph{Classifier Architecture and Training Objective}\label{sec:classifier}
We employ a lightweight feed-forward network to map feature vectors $\mathbf{z}_i$ to hallucination labels:
\begin{equation}
p_i = h_\theta(\mathbf{z}_i) = \sigma\left(\mathbf{w}_2^\top \phi(\mathbf{W}_1\mathbf{z}_i + \mathbf{b}_1) + b_2\right) \in (0,1)
\end{equation}
where $\phi$ denotes the ReLU activation function, $\sigma$ is the sigmoid function, and $\theta = \{\mathbf{W}_1, \mathbf{b}_1, \mathbf{w}_2, b_2\}$ are learnable parameters. Binary predictions follow $\hat{y}_i = \mathbf{1}\{p_i > \tau\}$ with threshold $\tau$ selected via validation set optimization.

The training objective employs standard binary cross-entropy loss:
\begin{equation}
\mathcal{L} = -\frac{1}{N}\sum_{i=1}^N \left[ y_i\log p_i + (1-y_i)\log(1-p_i) \right]
\end{equation}

We train model-specific classifiers to account for architectural differences in internal representations across LLM families. Optional contextual features (e.g., available function specifications) can be incorporated via the projection layer $\Pi$. Post-training calibration using temperature scaling is applied on held-out validation data to ensure well-calibrated probability estimates.

\begin{table*}[!h]
\centering
\caption{Comparison with NCP and Semantic Similarity baselines on tool-calling hallucination detection. Our method achieves competitive performance while requiring only a single forward pass, compared to multiple sampling approach required for NCP and Semantic Similarity. Results show hallucination class performance (Precision/Recall/F1-Score/Accuracy).}
\label{tab:baseline_comparison}
\begin{tabular}{llcccc}
\toprule
Model & Method & Precision & Recall & F1-Score & Accuracy \\
\midrule
\multirow{3}{*}{GPT-OSS-20B} & \textbf{Our Method} & \textbf{0.86} & \textbf{0.86} & \textbf{0.85} & \textbf{0.86} \\
\cmidrule{2-6}
& NCP & \textit{1.0} & \textit{0.79} & \textit{0.88} & \textit{0.95} \\
& Semantic Similarity & \textit{1.0} & \textit{0.79} & \textit{0.88} & \textit{0.95} \\
\midrule
\multirow{3}{*}{Llama-3.1-8B} & \textbf{Our Method} & \textbf{0.73} & \textbf{0.73} & \textbf{0.72} & \textbf{0.73} \\
\cmidrule{2-6}
& NCP & \textit{1.0} & \textit{0.733} & \textit{0.846} & \textit{0.935} \\
& Semantic Similarity & \textit{1.0} & \textit{0.731} & \textit{0.845} & \textit{0.935} \\
\midrule
\multirow{3}{*}{Qwen-7B} & \textbf{Our Method} & \textbf{0.81} & \textbf{ 0.74 } & \textbf{0.77} & \textbf{0.74} \\
\cmidrule{2-6}
& NCP & \textit{0.99} & \textit{0.45} & \textit{0.62} & \textit{0.94} \\
& Semantic Similarity & \textit{1.0} & \textit{0.44} & \textit{0.62} & \textit{0.94} \\
\bottomrule
\end{tabular}
\end{table*}

\subsection{Inference Protocol}\label{sec:inference}
Our method operates inline with generation, reusing the \emph{identical forward pass} that produces the tool call:

\begin{enumerate}
\item The LLM generates tool call $\tilde{f}(\boldsymbol{a})$ through standard autoregressive generation
\item Extract token positions $t^{\text{func}}$, $\mathcal{T}_{\text{args}}$, and $t^{\text{end}}$ from the final layer of this pass
\item Construct feature vector $\mathbf{z}$ according to Equation (1)
\item Compute hallucination probability $p = h_\theta(\mathbf{z})$ using the trained classifier
\item If $p > \tau$, execute predefined policy (execution blocking, user confirmation, fallback selection, or repair via re-prompting); otherwise proceed with normal tool execution
\end{enumerate}

Since feature extraction requires only final-layer representations from a single forward pass, computational overhead remains negligible, enabling deployment in latency-sensitive real-time agent systems. All experimental evaluations in Section~\ref{sec:experiments} adhere to this inference protocol.

\section{Experiments}\label{sec:experiments}

\subsection{Experimental Setup}

\subsubsection{Target Models}
We evaluate our approach on three representative open-source LLMs with diverse architectural characteristics:
\begin{itemize}
\item \textbf{Qwen7B}: A 7-billion parameter model employing Group Query Attention and RMSNorm, representing the Qwen model family's architectural innovations.
\item \textbf{GPT-OSS-20B}: A 20-billion parameter model following the standard transformer architecture with multi-head attention and LayerNorm.
\item \textbf{Llama-3.1-8B}: An 8-billion parameter model from the Llama family, incorporating RoPE positional embeddings and SwiGLU activation functions.
\end{itemize}
This selection provides coverage across different scales, architectural choices, and training methodologies to assess the generalizability of our approach.

\subsubsection{Baseline Methods}
We compare against Non Contradiction Probability (NCP)~\cite{hou2025probabilistic} and semantic similarity \cite{kuhn2023semantic}:
\begin{itemize}
\item \textbf{NCP} is the likelihood that a language model’s output does not conflict with verified facts or logical premises. To calculate NCP, we prompt the agent with the same test sample multiple times (n=3) and measure consistency using agreement of function name and parameter.
\item \textbf{Semantic Similarity} is another consistency technique measured using cosine similarity of responses from the agent over multiple invocations (n=3). Semantic similarity is calculated over the extracted tool call and parameters. 
\end{itemize}
Both baselines require multiple forward passes (5× computational overhead) compared to our single-pass approach.

\subsubsection{Dataset Construction}
Our evaluation focuses on mathematical reasoning domains where tool-calling precision is critical. We developed five specialized AI agents using the Glaive dataset~\cite{glaive2024} as the foundation:

\textbf{Agent Categories:}
\begin{itemize}
\item \textbf{Quick Calculator}: Basic arithmetic operations, unit conversions, and mathematical computations
\item \textbf{Personal Finance Suite}: Investment calculations, loan computations, and financial planning tools
\item \textbf{Health Assistant}: BMI calculations, caloric needs estimation, and health metric tracking
\item \textbf{Sustainability Assistant}: Carbon footprint calculations, energy efficiency metrics, and environmental impact assessments
\item \textbf{Digital Commerce Assistant}: Price calculations, discount computations, and transaction processing
\end{itemize}

\textbf{Data Processing:} From the Glaive dataset's multi-turn conversations, we extracted 2,411 tool-calling instances per model. Each instance contains: (1) user query, (2) contextual information, (3) ground-truth function call, and (4) predicted function call with cached final-layer representations. Tool names and parameters were canonicalized to ensure consistent evaluation across different syntactic representations.

\subsubsection{Training Configuration}
For each target model, we train a dedicated two-layer MLP classifier with model-specific input dimensions: 3584 for Qwen7B, 2880 for GPT-OSS-20B, and 4096 for Llama-3.1-8B. The classifier architecture consists of a hidden layer with 512 units and ReLU activation, followed by a single sigmoid output unit for binary classification. We apply dropout with probability 0.1 to the hidden layer during training to prevent overfitting.

We optimize the classifier using AdamW with a learning rate of $1 \times 10^{-4}$ and weight decay of $1 \times 10^{-5}$. Training proceeds for up to 50 epochs with early stopping (patience=5) based on validation loss. We use a batch size of 32 samples and employ cosine annealing with warm restarts for the learning rate schedule to improve convergence stability.

Our evaluation protocol follows a 60\%/20\%/20\% train/validation/test split. The decision threshold $\tau$ is selected via grid search over the range [0.1, 0.9] with step size 0.05, optimizing for balanced performance on the validation set. Post-training calibration using temperature scaling is applied on the held-out validation set to ensure well-calibrated probability estimates. We report standard classification metrics including precision, recall, F1-score, and accuracy for both classes, along with macro and weighted averages to account for class imbalance effects.

\begin{table*}[!t]
\centering
\caption{Detailed classification results for GPT-OSS-20B, Llama-3.1-8B and Qwen7-B using our final-layer feature extraction approach. }
\label{tab:main_results}
\begin{tabular}{llcccc}
\toprule
Model & Class & Precision & Recall & F1-Score & Support \\
\midrule
\multirow{5}{*}{GPT-OSS-20B} & Non-Hall. & 0.87 & 0.97 & 0.92 & 1824 \\
& Hall. & 0.86 & 0.53 & 0.66 & 587 \\
\cmidrule{2-6}
& Macro avg & 0.86 & 0.75 & 0.79 & 2411 \\
& Weighted avg & 0.86 & 0.86 & 0.85 & 2411 \\
\cmidrule{2-6}
& \textit{Accuracy} & \multicolumn{4}{c}{\textit{0.86}} \\
\midrule
\multirow{5}{*}{Llama-3.1-8B} & Non-Hall. & 0.73 & 0.82 & 0.77 & 1375 \\
& Hall. & 0.71 & 0.61 & 0.66 & 1036 \\
\cmidrule{2-6}
& Macro avg & 0.72 & 0.71 & 0.71 & 2411 \\
& Weighted avg & 0.73 & 0.73 & 0.72 & 2411 \\
\cmidrule{2-6}
& \textit{Accuracy} & \multicolumn{4}{c}{\textit{0.73}} \\
\midrule
\multirow{5}{*}{Qwen-7B} & Non-Hall. & 0.91 & 0.76      &  0.83      &  1667  \\
& Hall. & 0.34       &  0.62   &   0.44         & 333 \\
\cmidrule{2-6}
& Macro avg & 0.63        &  0.69      &0.64      & 2000 \\
& Weighted avg & 0.81         &  0.74     &  0.77        & 2000 \\
\cmidrule{2-6}
& \textit{Accuracy} & \multicolumn{4}{c}{\textit{ 0.74 }} \\
\bottomrule
\end{tabular}
\end{table*}

\subsection{Baseline Comparison}
We compare our approach against NCP and semantic similarity, methods for detecting hallucinations in black-box LLMs. Both the baselines operate on the principle that consistent facts align across multiple samplings, while hallucinated facts diverge. We adapt this approach to tool-calling scenarios by generating multiple tool call samples and measuring consistency.

The results show that while non-contradiction probability and semantic similarity baselines achieve higher precision and overall accuracy, our method based on internal LLM states provides a competitive and robust approach to hallucination detection with higher recall over the baselines across all models—especially noteworthy for GPT-OSS-20B, where recall is highest among all methods (0.86 vs. 0.74 for Qwen-7B and 0.73 for Llama-3.1-8B), showing greater sensitivity to diverse hallucination types.

Internal state methods enable inference-time hallucination detection, operating directly within the LLM’s decoding process, which eliminates the need for expensive sampling or repeated passes over the output. Unlike external-sampling such as NCP and semantic similarity or post-hoc semantic scoring, our method leverages information that is only available during stepwise generation—unlocking model-specific heuristics unreachable by black-box techniques. This makes our technique more adaptable for streaming, edge, and low-latency production scenarios, where fast and actionable hallucination risk scores are needed.

\subsection{Main Results}
Table~\ref{tab:main_results} presents our approach's performance on GPT-OSS-20B, Llama-3.1-8B and Qwen-7B, which demonstrate the effectiveness of internal representation-based hallucination detection.

GPT-OSS-20B achieves strong overall performance with 86\% accuracy and demonstrates effective hallucination detection with 86\% precision and 53\% recall. The model maintains high precision in non-hallucination detection (87\%) while achieving balanced macro-averaged performance (F1=0.79). Llama-3.1-8B shows more balanced performance across both classes, with nearly equal class distribution (1375:1036 ratio) and consistent precision (73\% vs 71\%) for both hallucination and non-hallucination detection.

The results demonstrate that internal transformer representations contain sufficient information to distinguish between correct and hallucinated tool calls. The computational efficiency of our approach—requiring only final-layer representations from a single forward pass—makes it suitable for real-time deployment in production agent systems.

\begin{table*}[h]
\centering
\caption{Ablation study on feature extraction methods for hallucination detection using Qwen-7B.}
\label{tab:ablation_features}
\resizebox{1.5\columnwidth}{!}{%
\begin{tabular}{lcccccc}
\toprule
\textbf{Method} & \textbf{Acc.} & \textbf{AUC} & \textbf{Prec.} & \textbf{F1} & \textbf{Dim.} & \textbf{Time Complexity} \\
\midrule
Last-Layer Rep. & \textbf{0.746} & \textbf{0.721} & \textbf{0.749} & \textbf{0.748} & $d$ & $O(n \cdot d)$ \\
Statistical & 0.745 & 0.719 & 0.748 & 0.746 & $2d$ & $O(n \cdot d)$ \\
Attention Weighted & 0.744 & 0.719 & 0.748 & 0.746 & $d$ & $O(n^2 \cdot d)$ \\
Last Token & 0.744 & 0.720 & 0.747 & 0.745 & $d$ & $O(1)$ \\
First-Last-Mean & 0.743 & 0.719 & 0.747 & 0.744 & $3d$ & $O(n \cdot d)$ \\
Statistical Ext. & 0.743 & 0.718 & 0.746 & 0.744 & $4d$ & $O(n \cdot d)$ \\
CLS Token & 0.742 & 0.719 & 0.746 & 0.744 & $d$ & $O(1)$ \\
Multi-Scale & 0.742 & 0.718 & 0.745 & 0.743 & $5d$ & $O(n \cdot d)$ \\
Max Pooling & 0.741 & 0.717 & 0.744 & 0.742 & $d$ & $O(n \cdot d)$ \\
Min Pooling & 0.739 & 0.716 & 0.743 & 0.741 & $d$ & $O(n \cdot d)$ \\
\bottomrule
\end{tabular}%
}
\end{table*}

\subsection{Ablation Study on Feature Extraction Methods}

To understand the impact of different feature extraction strategies on hallucination detection performance, we conduct a comprehensive ablation study examining various approaches to aggregate transformer hidden states into fixed-size representations for classification. We conduct all of the experiments on Qwen-7B.

\subsubsection{Feature Extraction Methods}

We evaluate ten distinct feature extraction methods, each representing different strategies for converting variable-length sequence representations into fixed-size feature vectors:

\noindent\textbf{Basic Pooling Methods:}
\begin{itemize}
    \item \textbf{Last-Layer Representations}: Mean pooling across the sequence dimension of the final transformer layer, producing a fixed-size representation equal to the model's hidden dimension.
    \item \textbf{Max Pooling}: Element-wise maximum across the sequence dimension, capturing the most activated features for each dimension.
    \item \textbf{Min Pooling}: Element-wise minimum across the sequence dimension, identifying consistently low-activated features.
\end{itemize}

\textbf{Token-Specific Methods:}
\begin{itemize}
    \item \textbf{CLS Token}: Uses only the [CLS] token representation from the final layer, leveraging the token specifically designed for classification tasks.
    \item \textbf{Last Token}: Extracts features from the final token in the sequence, capturing end-of-sequence information.
\end{itemize}

\textbf{Statistical Aggregation Methods:}
\begin{itemize}
    \item \textbf{Statistical}: Concatenates mean and standard deviation across the sequence dimension, doubling the feature dimensionality ($2d$ features).
    \item \textbf{Statistical Extended}: Combines mean, standard deviation, minimum, and maximum statistics, quadrupling the feature dimensionality ($4d$ features).
\end{itemize}

\textbf{Advanced Aggregation Methods:}
\begin{itemize}
    \item \textbf{Attention-Weighted Pooling}: Computes attention weights based on L2 norms and performs weighted averaging across the sequence.
    \item \textbf{First-Last-Mean}: Concatenates the first token, last token, and mean-pooled representations, tripling the feature dimensionality ($3d$ features).
    \item \textbf{Multi-Scale}: Combines multiple pooling strategies (mean, max) with positional tokens (first, middle, last), creating a comprehensive representation ($5d$ features).
\end{itemize}

We evaluate each feature extraction method on the Qwen-7B using the digital commerce assistant dataset. All experiments use identical training configurations, including the same train-test split (70-30), optimization parameters, and evaluation metrics to ensure fair comparison.

Our ablation study reveals several key insights for hallucination detection in language models:

\begin{enumerate}
    \item \textbf{Simple aggregation methods are effective}: Mean pooling across the sequence dimension provides the best balance of performance and simplicity.
    
    \item \textbf{Diminishing returns of complexity}: More sophisticated aggregation methods with higher dimensionality do not necessarily yield better performance, suggesting that the base transformer representations already capture the essential information.
    
    \item \textbf{Sequence-level information is valuable}: Methods that aggregate information across the entire sequence consistently outperform single-token approaches, indicating that hallucination signals are distributed throughout the generated text.
    
    \item \textbf{Robustness across methods}: The relatively small performance variance across methods suggests that the underlying transformer representations are robust and that multiple aggregation strategies can effectively capture hallucination patterns.
\end{enumerate}

Based on these findings, we recommend using last-layer representations (mean pooling) as the default feature extraction method for transformer-based hallucination detection, as it provides optimal performance while maintaining computational efficiency and implementation simplicity.

\section{Conclusion and Limitations}
In this paper, we presented a novel approach for detecting tool-calling hallucinations in LLMs by leveraging their internal representations, demonstrating strong performance across multiple models with accuracies ranging from 72.7\% to 86.4\%. Our method's effectiveness is particularly notable in GPT-OSS-20B, which achieved 86\% precision in hallucination detection while maintaining balanced performance across classes. The approach's ability to operate in real-time using only last-layer representations from a single forward pass makes it practically viable for deployment in production environments. Future work will investigate the possibility of developing a unified hallucination detector that works effectively across different model families, potentially identifying common patterns in internal representations that generalize across architectures.

However, reference-agreement labeling inherits problems from dataset calls and standardization rules, and determining function equivalence beyond string matching remains difficult without additional context. While our three-point feature design is intentionally simple, future work could explore field-aware pooling or lightweight attention over argument spans while keeping single-pass requirements, alongside expanded experiments, accuracy analysis, and robustness studies across new tools.

\bibliography{aaai2026}

\end{document}